\documentclass[letterpaper]{article} 
\usepackage[]{aaai23} 
\usepackage{times} 
\usepackage{helvet}
\usepackage{courier}
\usepackage[hyphens]{url}
\usepackage{graphicx}
\usepackage{natbib} 
\usepackage{caption}
\usepackage{array}
\usepackage{booktabs}
\usepackage{amssymb}
\usepackage{mathtools}
\usepackage[subtle]{savetrees}
\usepackage[table]{xcolor} 
\usepackage{algorithm}
\usepackage{algorithmic}
\usepackage{newfloat}
\usepackage{listings}
\DeclareCaptionStyle{ruled}{labelfont=normalfont,labelsep=colon,strut=off} 
\newcolumntype{L}{>{\centering\arraybackslash}m{5cm}}

\floatstyle{ruled}
\newfloat{listing}{tb}{lst}{}
\floatname{listing}{Listing}

\usepackage{hyperref}
\hypersetup{colorlinks=True}

\title{GIST: Generating Image-Specific Text for Fine-grained Object Classification}
\author{
    Kathleen M. Lewis\equalcontrib\textsuperscript{\rm 1},
    Emily Mu\equalcontrib\textsuperscript{\rm 1},
    Adrian V. Dalca\textsuperscript{\rm 1,2,3},
    John Guttag\textsuperscript{\rm 1}
}
\affiliations{
    \textsuperscript{\rm 1}Massachusetts Institute of Technology\\
    \textsuperscript{\rm 2}
    Harvard Medical School\\
    \textsuperscript{\rm 3} Massachusetts General Hospital
}

\begin{document}

\maketitle

\begin{abstract}
Recent vision-language models outperform vision-only models on many image classification tasks. However, because of the absence of paired text/image descriptions, it remains difficult to fine-tune these models for fine-grained image classification. In this work, we propose a method, GIST, for generating \textit{image-specific} \textit{fine-grained} text descriptions from image-only datasets, and show that these text descriptions can be used to improve classification. Key parts of our method include 1. prompting a pretrained large language model with \textit{domain-specific} prompts to generate diverse fine-grained text descriptions for each class and 2. using a pretrained vision-language model to match each image to label-preserving text descriptions that capture relevant visual features in the image. We demonstrate the utility of GIST by fine-tuning vision-language models on the image-and-generated-text pairs to learn an aligned vision-language representation space for improved classification. We evaluate our learned representation space in full-shot and few-shot scenarios across four diverse fine-grained classification datasets, each from a different domain. Our method achieves an average improvement of $4.1\%$ in accuracy over CLIP linear probes and an average of $1.1\%$ improvement in accuracy over the previous state-of-the-art image-text classification method on the full-shot datasets. Our method achieves similar improvements across few-shot regimes. Code is available at \href{https://github.com/emu1729/GIST}{https://github.com/emu1729/GIST}.
\end{abstract}

\section{Introduction}
\begin{figure}[tb]
\begin{center}
\centerline{\includegraphics[width=\linewidth]{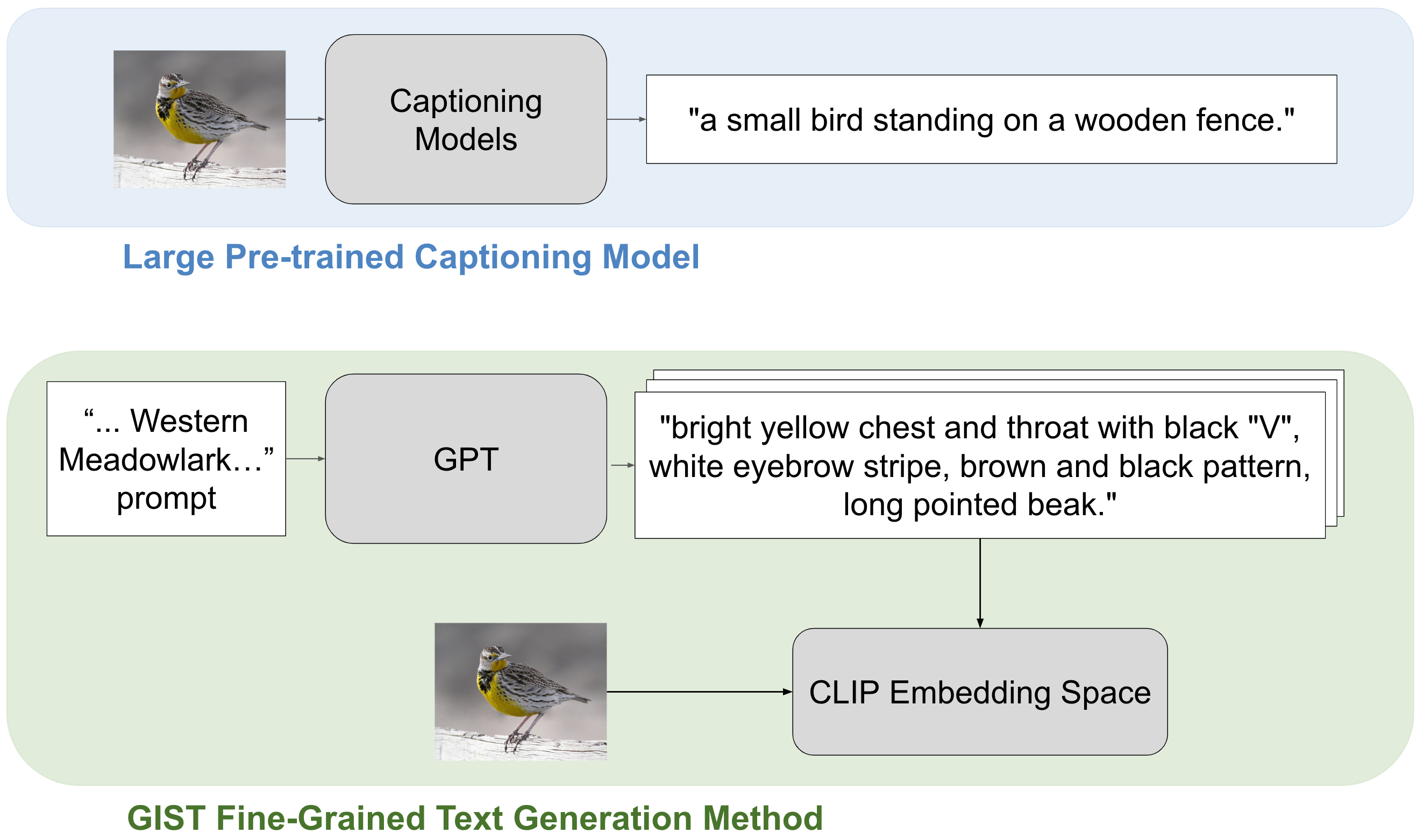}}
\caption{Large pretrained vision-language models (top), such as GIT, perform well on tasks at the coarse-object-level, but don't always generalize well to fine-grained tasks. Our method (bottom), GIST, generates fine-grained class-specific text descriptions. Our method then uses CLIP to match each image to the closest label-preserving text. This results in fine-grained image-text pairs where the text describes specific visual features in the image.
}
\label{fig:placeholder}
\vspace{-10mm}
\end{center}
\end{figure}

The recent development of foundation models pretrained on image-text pairs has led to impressive performance across vision and language tasks such as zero-shot classification, image generation, and image captioning~\cite{radford2021learning,DALLE,GIT}. Fine-tuning these models for domain-specific tasks, however, requires image-text pairs, which can be costly to obtain~\citep{zhou2022conditional, zhou2022learning}. In particular, fine-tuning these models for fine-grained image classification requires the construction of image-specific text prompts that differentiate visual features between sub-categories of objects, e.g., species of birds. These fine-grained captions can be difficult to obtain. Pretrained image captioning methods, such as GIT~\cite{GIT}, can generate descriptive captions for coarse-object-level interactions, but are not able to generate the details needed for fine-grained classification (Figure~\ref{fig:placeholder}).

In contrast, pretrained large language models, such as GPT~\cite{gpt,openai2023gpt4}, contain rich prior knowledge about both coarse and fine-grained classes. Large language models are able to generate fine-grained text descriptions when prompted about a specific class; however, these descriptions are not specific to a particular image. Recent works have explored how to bridge this gap to use text generated by GPT to improve few-shot image classification~\cite{LaBo,prompt2023}. These works use generic prompt templates, such as ``What does a \textit{cardinal} look like?'', to generate class-specific text descriptions. We show that instead of using generic templates, constructing \textit{domain-specific} prompts that highlight potential sub-class differences can lead to more specific text descriptions that differentiate visual features of fine-grained classes and lead to better classification performance. By matching each training image to specific captions, we find that we can construct diverse image text pairs to fine tune multi-modal networks for fine-grained classification tasks.

We introduce GIST (Generating Image-Specific Text), a streamlined approach for generating fine-grained, image-specific text descriptions for \textit{any-shot} fine-grained classification. We use GPT to first generate descriptive and diverse class-specific captions to cover the possible range of visual appearances of each class by constructing dataset-specific prompts. For example, we might prompt GPT with ``What does a \textit{male} Western Meadowlark look like?'' or ``What does becker nevus look like on the \textit{shoulder}.'' for classes in the bird or dermatology domains respectively. We then use CLIP to match each training image to the closest label-preserving text descriptions conditioned on training label. With this process, our method can generate a large variety of possible visual descriptions and then match each image to the text descriptions that are specific to the image content. After automatically generating and matching text descriptions to each training image, our method uses GPT to summarize the text descriptions into a concise format well suited for fine tuning CLIP. We use contrastive learning to fine-tune CLIP and learn an aligned vision-language representation space for classification.

We evaluate our method against recent vision-language classification methods~\cite{LaBo, prompt2023, Goyal2023,radford2021learning} and image-only classification methods~\cite{ResNet} on four fine-grained datasets from diverse domains: dermatology~\cite{groh2021evaluating}, flowers~\cite{flowers102},  birds~\cite{CUB2011}, and aircraft~\cite{fgvc}. We show that our model achieves state-of-the-art results on all four datasets across many different training scenarios. Unlike other methods that are optimized for either full or few-shot classification, our method is consistently better than the state-of-the-art on both.

The key contributions in our paper are:
\begin{itemize}
    \item We introduce GIST, a method for generating fine-grained, image-specific text. To our knowledge, we are the first to use language priors and image-text foundation models to generate image-specific captions for fine-grained datasets.
    \item We demonstrate that fine-tuning CLIP on our GIST image-text pairs outperforms recent vision-language classification methods on four fine-grained datasets for full-shot and few-shot classification regimes. 
    \item We compare our CLIP fine-tuning approach to a visual-grounding approach to analyze how to best use the generated image-specific text descriptions.
    \item We show that fine-tuning with GIST-generated text improves upon baselines for all CLIP models and study the effects of the number of captions per image, and caption length on classification performance. 
    \item We provide a new fine-grained dermatology dataset, Fitzpatrick40.
\end{itemize}

\section{Related Work}

\subsection{Vision-Language Models}
Models pretrained on large-scale multimodal datasets have been shown to provide data representations that are transferable to many tasks and domains \citep{joulin2016learning, li2017learning, desai2021virtex, sariyildiz2020learning}. CLIP~\cite{radford2021learning} is pretrained using contrastive learning on a dataset of 400 million image-text pairs and has shown strong performance for both zero-shot learning and other downstream tasks. Other related approaches achieve similar performance with larger and noisier datasets \citep{jia2021scaling}, or fine-tune CLIP to adapt it to out-of-domain datasets \citep{kumar2022fine,miller2021accuracy, wortsman2022robust,zhai2022lit}. Still other methods have tried to improve the efficacy of CLIP training by incorporating self-supervised training techniques \citep{mu2022slip,li2021supervision,li2023scaling}, re-writing captions for pretraining images \citep{fan2023improving}, and generating captions for images using off-the-shelf captioning models \citep{nguyen2023improving}. We show in our ablation experiments that GIST captions contain more specific fine-grained information than off-the-shelf generated captions and result in better classification performance. 

Large-scale captioning models~\cite{GIT,Coca} leverage image-text pairs to learn a shared representation space that is useful for generating text from images. These visual grounding methods are useful for generating high-level descriptions of objects. However, we show in our experiments that the captioning models do not perform well on fine-grained classes. Large-scale image generation models such as DALL-E~\cite{DALLE}, Imagen~\cite{saharia2022photorealistic}, and stable diffusion~\cite{rombach2022high} are trained on large-scale image-text data in order to learn to generate photo-realistic images from text. In the few-shot learning literature, DALL-E has been shown to help by augmenting small image datasets with synthetic images\cite{prompt2023}. This works well for Imagenet \citep{fan2023improving} classes, but we show in our experiments the generated images are not specific or accurate enough for fine-grained domains, such as dermatology diseases.

\subsection{Large Language Models}
Recently, large language models trained with huge quantities of textual data have achieved remarkable human-level performance on many NLP tasks \citep{BERT,biobert,gpt, touvron2023llama,chowdhery2022palm}. GPT~\cite{gpt, openai2023gpt4} is a self-supervised pretrained large language model with 175 billion parameters. GPT has knowledge across many domains and can be used for a variety of language tasks, including text generation. GPT can generate accurate, detailed descriptions for many different domains, especially when it is first prompted with structured examples of the desired text format. We use the strong prior knowledge of GPT to generate text descriptions and show that these text descriptions carry more useful information than generated image captions from models such as GIT. We further show that matching these captions to images and using image-text pairs for fine-tuning CLIP improves classification over image-only methods, suggesting that GPT-generated descriptions hold useful knowledge about fine-grained classes. 

 \begin{figure*}[tb]
\begin{center}
\centerline{\includegraphics[width=0.9\linewidth]{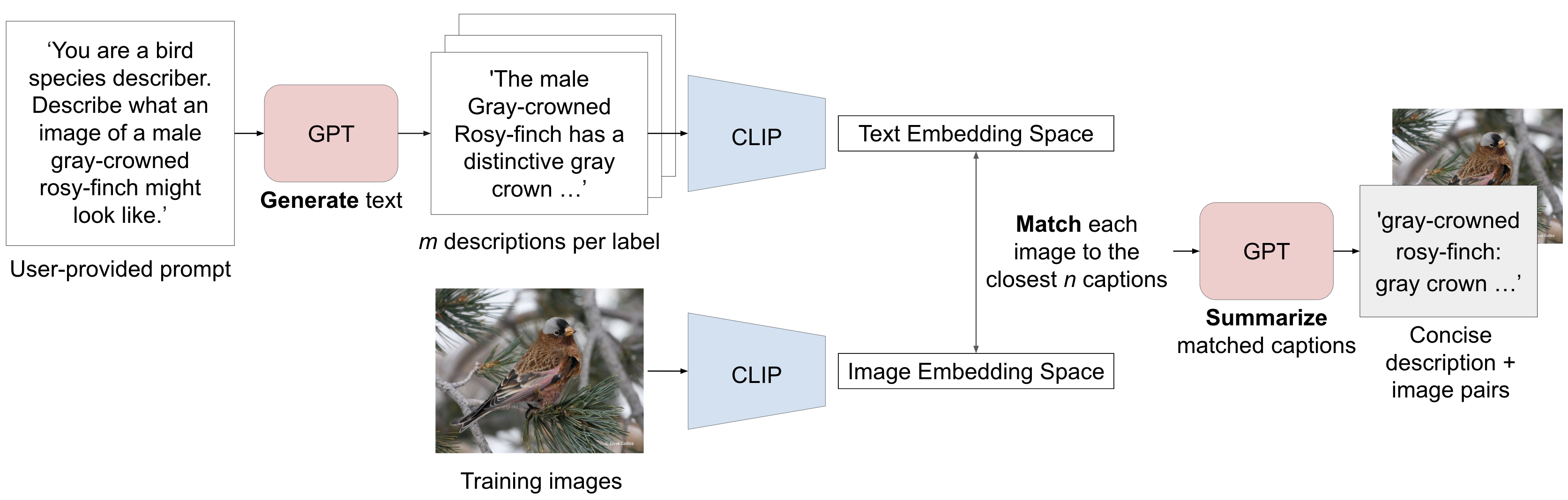}}
\caption{An overview of the GIST method. We leverage prior domain knowledge by using GPT to generate multiple text descriptions for each label. We use CLIP to match each training image to the closest \textit{n} GPT descriptions with the same label. We then use GPT to summarize each of the matched captions.}
\label{fig:textGeneration}
\vspace{-7mm}
\end{center}
\end{figure*}

\subsection{Few-shot Classification using Foundation Models}
There are several existing methods that leverage foundation models for improving few-shot classification. CLIP~\cite{radford2021learning} presents classification results from using zero-shot and few-shot linear-probe methods, which have become common baselines to compare to in the vision-language few-shot learning literature. Since CLIP is a pretrained model with an aligned image-text representation space, zero-shot classification can be performed by encoding class names in the form ``a photo of a...'' and then finding the closest class embedding to each encoded image. The CLIP linear-probe classification method involves learning a linear classifier on the frozen CLIP image embeddings. We show in our experiments that fine-tuning CLIP on our image and generated text pairs improves classification accuracy over zero-shot and CLIP linear-probing.

LaBo~\cite{LaBo} uses GPT-3 to generate candidate concepts for each class and aligns these text concepts to the images using CLIP. Our results demonstrate that our image-text description matching and use of contrastive learning to align the image and text embeddings improves over their approach. The Prompt, Generate, then Cache method~\cite{prompt2023} use foundation models to generate text and synthetic images to supplement few-shot learning. They then use CLIP and contrastive learning to ensemble predictions. Another recent method~\cite{udandarao2022sus} also generates images to improve upon zero-shot CLIP performance. We show in our results that synthetic images do not generalize well to fine-grained domains, hindering performance. FLYP~\cite{Goyal2023} uses class names as text descriptions and contrastive loss to align image-text embeddings. Another recent method~\cite{Lin2023} also uses class names to construct text prompts and trains a linear classifier on the shared image-text representation space. We show that generating full text descriptions and matching descriptions to images outperforms using only the class name. ProD~\cite{Tianyi2023} trains on multiple datasets to learn both domain-specific prompts from image features and a domain-general prompt that improves cross-domain generalization. Their experiments demonstrate that having both domain-specific and domain-independent prompts improves few-shot classification accuracy over a ``CNN+Transformer'' baseline. There is no public code available for ProD. A concurrent paper~\cite{Maniparambil2023} to our work uses GPT-4 and a pre-trained CLIP model to improve zero-shot and few-shot classification.

\section{Method} 
\begin{figure}[tb]
\begin{center}
\centerline{\includegraphics[width=\linewidth]{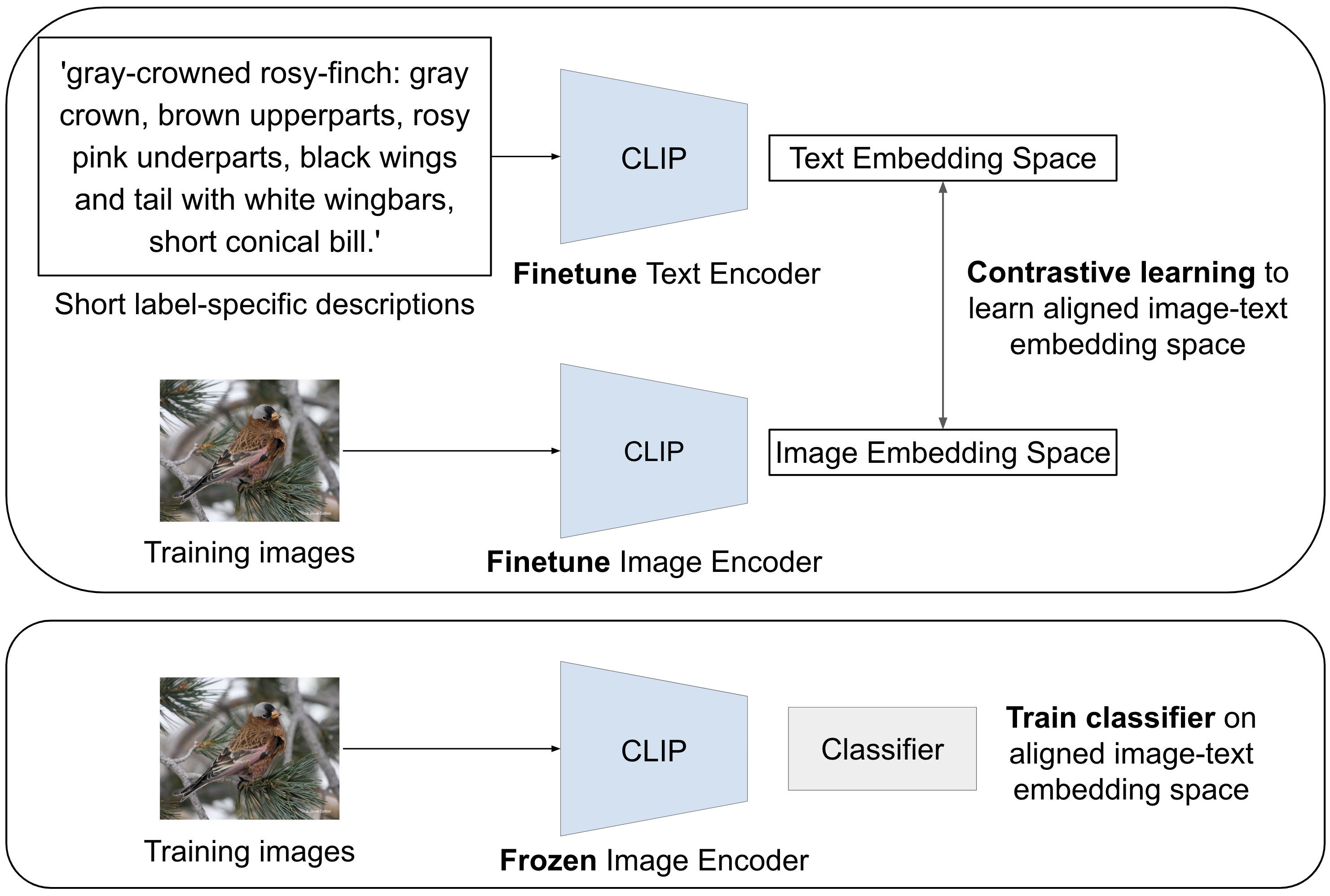}}
\caption{An overview of our contrastive fine-tuning and classification methods. We fine-tune CLIP using our paired training images and generated image-specific captions. Once we have learned an aligned image-text embedding space, we freeze the CLIP encoders and train a classifier.}
\label{fig:method}
\vspace{-7mm}
\end{center}
\end{figure}

GIST is designed for fine-grained image tasks, where image labels are subcategories of a broader category. In this section, we first describe the GIST method for generating image-specific text descriptions (Figure~\ref{fig:textGeneration}) and then present how to apply GIST to fine-grained image classification (Figure~\ref{fig:method}).

\subsection{Problem Formulation} 
Consider a set of images $i \in I$ that each have a fine-grained label mapping the image to a class $y \in Y$. Our goal is to generate fine-grained text, $t_i$, for each image. Suppose we have a pretrained image-text alignment model (e.g. CLIP \citep{radford2021learning}), where the quality of image-text alignment can vary depending on domain. Let $T$ represent the space of text descriptions for images. This pretrained image-text alignment model consists of an image encoder $f_I : I \mapsto \mathbb{R}^d$ and text encoder $f_T : T \mapsto \mathbb{R}^d$. Suppose that we have a pretrained large language model, $L$, from which we can generate sets of class-specific descriptions based on language priors $D_y = L(prompt(y))$.

\subsection{Generating Image-Specific Text} We use large language models to supplement our image dataset with text. For each class $y \in Y$, we generate \textit{m} class-specific captions, where \textit{m} depends on domain and language priors $D_y = L(prompt(y))$. We then use our pretrained image-text alignment model to match each image in our training set $i \in I$ with our generated class-specific text to get image-specific text, $t_i$. We show an overview of the GIST method in Figure~\ref{fig:textGeneration}.
\subsubsection{Prompt Construction}Previous methods have prompted large language models with generic templates, such as ``What does a \textit{Blue Jay} look like?''~\cite{LaBo,prompt2023}. However, we show that when generating sample captions for fine-grained classification problems, it is important to construct prompts that include \textit{domain-specific} characteristics. These domain-specific prompts help significantly for realistic downstream caption matching to images. For example, prompting GPT-3 while iterating through a list of different body parts and a disease label (e.g. ``What does Behcet's disease look like on the mouth?'' rather than ``What does Behcet's disease look like?'') gives us more realistic captions than just using the disease label alone. We provide a list of the prompts used for each of the domains in our experiments in the supplement. Note that we do not modify the prompting schemes between specific labels but only for entire domains. We find that careful thought upfront about potential domain fine-grained class differences (e.g. male bird versus female bird appearances) results in better captions. This prompt construction requires some manual input from a user, but is not time intensive and does not require detailed domain knowledge. For example, one might need to know that dermatology diseases can appear on different parts of the body, but not which diseases appear where.  \subsubsection{Image-specific Text Generation} For each image $i$ and corresponding label $y$, we use the pretrained image and text encoders to compute embeddings $f_I(i)$ and $f_T(d)$ for each class-specific description $d \in D_y$. Given these embeddings, we automatically select the $n$ most similar description embeddings in $D_y$ to $f_I(i)$ to match each image in our training set. We determine embedding similarity by computing the cosine similarity over the L2 normalized embeddings. Finally, we use the pretrained large language model, $L$, to summarize the matched text descriptions into a more concise form. Following summarization, we append the class name to the concise caption. We find that performing contrastive fine-tuning with concise captions improves downstream classification performance relative to long captions. We show a summary of the text generation process in Figure~\ref{fig:textGeneration} and include example long and concise text prompts in Section~\ref{sec:experiments} and in the supplement. 

\subsection{Contrastive Fine-tuning for Image Classification} We find that contrastive fine-tuning improves fine-grained image classification over using a linear-probe on frozen pretrained image embeddings. Following our text description generation method, we have training set image $i$ and an associated set of generated text descriptions $t_i$, where $|t_i| = n$. We use both the images and generated text descriptions to fine tune the image and text encoders to align the images and generated text as shown in Figure~\ref{fig:method}. Concretely, we optimize the fine-tuning objective,

\vspace{-2.5mm}
\begin{align*}
L(D', \theta) &\coloneqq \sum_{i=1}^{B} -\text{log}\frac{\text{exp}(f_I(i_i)*f_T(t_i))} {\sum_{j=1}^{B}\text{exp}(f_I(i_i)*f_T(t_j))} + \\
& \sum_{i=1}^{B} -\text{log}\frac{\text{exp}(f_I(i_i)*f_T(t_i))}{\sum_{j=1}^{B}\text{exp}(f_I(i_j)*f_T(t_i))}
\end{align*}
for a batch of images and their generated text descriptions $D'={(i_0, t_0), ... }$, where $t_0 \in t_{i_0}$ and $\theta$ represents all of the image and text encoder parameters. Once we have aligned embeddings, we update the parameters in the image and text encoders $f'_I$ and $f'_T$. We then train a linear probe classifier on the images by first encoding images using $f'_I$ into the aligned text-image embedding space. 

\section{Experiments}\label{sec:experiments}

In this section, we provide implementation details and evaluate GIST on fine-grained classification. We first compare GIST to several recent state-of-the-art benchmarks on any-shot classification performance for four fine-grained classification datasets. We then evaluate all methods using different CLIP models. We further analyze how selecting GIST parameters, including number of matched captions and the length of matched captions, effects downstream classification performance. Finally, we perform a comparison of GIST against visual grounding and provide qualitative results.

\paragraph{Datasets.} We evaluate on four fine-grained classification datasets, \textbf{CUB200-2011}~\cite{CUB2011}, \textbf{Flowers102}~\cite{flowers102}, \textbf{FGVC-Aircraft}~\cite{fgvc}, and \textbf{Fitzpatrick40}, a cleaned subset of Fitzpatrick17k~\cite{groh2021evaluating}. The CUB200-2011 dataset has 11,788 labeled images of 200 different bird species. We use the published train/test split and separate 10\% of the training set as a validation set. The Flowers102 dataset has 8,189 labeled images of 102 different flower species. We use the published train/val/test split. The FGVC-Aircraft dataset has 10,000 labeled images of 100 different types of aircrafts. We use the published train/val/test split. The Fitzpatrick17k dataset has 16,577 labeled images for 114 different dermatology diseases. The dataset is known to have erroneous images~\cite{pakzad2023circle}, and we also find that it has many mislabeled images, where the image label in the dataset does not match the image label from the image's original medical website source. We clean a 40-label subset of the 114-label original dataset. We denote this cleaned dataset as Fitzpatrick40. We include more information about the dataset and the dataset cleaning process in the supplement. Fitzpatrick40 has 2,609 images, which we split into 2,115 training, 222 validation, and 272 test. We provide the cleaned Fitzpatrick40 dataset as an additional contribution of this work. For all four datasets, we generate \textit{k}-shot datasets by randomly sampling \textit{k} training images from each label for \textit{k}=1,3,5.

\paragraph{Implementation.} We use the ViT-L/14@336px CLIP model for all experiments. We also include a study on how different CLIP models impact fine-grained classification performance in Section~\ref{sec:method-analysis}. For the Fitzpatrick, Flowers102, and FGVC-Aircraft datasets, we generate text descriptions using GPT-3~\cite{gpt}. For the CUB200-2011 dataset, we generate text descriptions using GPT-4~\cite{openai2023gpt4}. We generate between 20 and 60 different class-specific captions, depending on the dataset. 

\paragraph{Baselines.} We compare our method to both image-only and multimodal classification baselines.
\paragraph{Image-only Baselines.}
\begin{itemize} 
\setlength{\parskip}{0pt}
\vspace{-3pt}
    \item \textbf{ResNet}~\cite{ResNet} We train a ResNet50 network for 600 epochs using standard SGD with a learning rate of 0.001, batch size of 64, momentum of 0.9, and 1e-4 weight decay. 
    \item \textbf{ResNet with Class Re-weighting (ResNet-RW)} We train the improved baseline with class re-weighting to compensate for the class imbalance across labels. We use a weighted random sampler in our data loader to sample from each class with equal probability.
    \end{itemize}
    \paragraph{Multimodal Baselines.}
    \begin{itemize} 
    \setlength{\parskip}{0pt}
    \vspace{-3pt}
    \item \textbf{Zero-shot CLIP (CLIP ZS)}~\cite{radford2021learning} Since CLIP models are trained to align image-text datasets, we perform zero-shot classification using the pretrained CLIP model. Given $y \in Y$ class names, we use standard templates (e.g. ``a photo of a ...'', ``a picture of a ...'') to construct text descriptions for each class $y$. We then take the average embedding $\overline{f_T}(y)$ of the templates as the class embedding. The zero-shot class prediction for image $i$ corresponds to the most similar class embedding to image embedding $f_I(i)$, where similarity is defined as cosine similarity of the l2-normalized embeddings.
    \item \textbf{Linear-probe CLIP (CLIP LP)}~\cite{radford2021learning} We learn a linear classifier on the frozen image embeddings computed using the pretrained image encoder $f_I$. This linear classifier is trained for 500 epochs using standard SGD with a learning rate of 0.05, batch size of 64, momentum of 0.9, and 1e-4 weight decay. 
    
    \item \textbf{Language in a Bottle (LaBo)}~\cite{LaBo} We train the LaBo method, using their open source code, on the full datasets and our \textit{k}-shot datasets. We evaluate their method on a suite of CLIP models and compare to ViT-L/14@336px. We generate text concepts using GPT-3 \citep{brown2020language} for the Fitzpatrick, FGVC, and Flowers102 datasets and using GPT-4 \citep{openai2023gpt4} for the CUB200-2011 dataset for a fair comparison to our approach. 
    \item \textbf{Cascade of Foundation Models (CaFo)}~\cite{prompt2023} We train the CaFo model, using their open source code, on our \textit{k}-shot datasets. We evaluate their method with several CLIP models and compare to the final results using ViT-L/14@336px. For the Flowers102 and FGVC datasets, we use their pre-generated DALL-E images and GPT-3 prompts. For the other two datasets, we generate our own DALL-E images and GPT prompts following the process from their paper. We use the same GPT versions as our method to stay consistent. For each k-shot experiment, we try using both 1 DALL-E generated image as well as k DALL-E images. We report the best score. We do not compare to the full training set results since these are not reported in the original paper, and their method was developed for few-shot learning.
    \item \textbf{Finetune Like You Pretrain (FLYP)}~\cite{prompt2023} We fine-tune the pretrained CLIP model using class names only with simple templates (e.g. ``a photo of a ...'') using contrastive training. We then train a linear probe on the learned image encoding and report the classification accuracy. 
    \end{itemize}

\begin{table*}[!htbp]
\centering
\small
\renewcommand{\arraystretch}{1.3} 
\rowcolors{2}{gray!25}{white}
\begin{tabular}{*9c}
\toprule
Methods &  \multicolumn{2}{c}{Full Dataset} & \multicolumn{2}{c}{5-shot} & \multicolumn{2}{c}{3-shot} & \multicolumn{2}{c}{1-shot}\\
{}   & \textit{Top-1}   & \textit{Top-3}   & \textit{Top-1}   & \textit{Top-3} & \textit{Top-1}   & \textit{Top-3} & \textit{Top-1}   & \textit{Top-3}\\
\midrule
ResNet   & 67.30 (2.87)  &  85.67 (2.12)  &  31.62 (3.69) & 50.37 (2.61) & 19.98 (1.73) & 40.32 (1.35) & 12.13 (1.97)  & 25.73 (0.79)  \\
ResNet-RW  & 70.95 (2.84) & 85.62 (2.16) & - & - & - & - & - & -  \\
CLIP ZS & 2.94 (1.06) & 7.40 (1.61) & 2.94 (1.06) & 7.40 (1.61) & 2.94 (1.06) & 7.40 (1.61) & 2.94 (1.06) & 7.40 (1.61) \\
CLIP LP & 71.19 (2.84) & 90.13 (1.76) & 41.49 (2.99) & 63.59 (3.01) & 36.07 (2.94) & 58.81 (2.99) & 18.77 (2.23) & 37.05 (2.88) \\
LaBo & 57.35 (0.19) & 81.62 (0.20) & 32.17 (0.25) & 60.47 (0.26) & 31.62 (0.08) & 54.78 (0.12) & 24.63 (1.04) & \textbf{53.68 (0.58)} \\
CaFo & - & - & 40.32 (0.63) & 65.93 (1.83) & 31.62 (2.67) & 57.12 (2.11) & \textbf{25.37 (1.08)} & 47.92 (1.21) \\
FLYP & 73.16 (2.70) & 89.73 (1.89) & 42.18 (2.98) & 65.50 (2.96) & 39.87 (3.00) & 61.20 (2.97) & 19.90 (2.37) & 38.35 (2.95) \\
\textbf{GIST (Ours)} & \textbf{75.77 (2.67)} & \textbf{90.81 (1.78)} & \textbf{46.96 (3.13)} & \textbf{71.29 (2.82)} & \textbf{41.24 (2.95)} & \textbf{63.60 (3.01)} & 21.06 (2.38) & 38.28 (2.91) \\ 
\bottomrule
\end{tabular}
\caption{Fine-tuning with GIST captions achieves best top-1 and top-3 accuracy for Fitzpatrick40 in comparison to baseline methods for full-shot, 5-shot, and 3-shot regimes. We report the average accuracy and standard deviation over three k-shot samples. For the full dataset and for the CLIP zero-shot experiments, we compute statistics over 1000 bootstrapped samples of the test set.}
\label{tab:fitz}
\end{table*}

\begin{table*}[!htbp]
\centering
\small
\renewcommand{\arraystretch}{1.3} 
\rowcolors{2}{gray!25}{white}
\begin{tabular}{*9c}
\toprule
Methods &  \multicolumn{2}{c}{Full Dataset} & \multicolumn{2}{c}{5-shot} & \multicolumn{2}{c}{3-shot} & \multicolumn{2}{c}{1-shot}\\

{}   & \textit{Top-1}   & \textit{Top-3}   & \textit{Top-1}   & \textit{Top-3} & \textit{Top-1}   & \textit{Top-3} & \textit{Top-1}   & \textit{Top-3}\\
\midrule
ResNet   & 79.08 (0.56) &  92.63 (0.35)  & 49.83 (0.47)  & 72.05 (0.24) & 34.89 (0.55) & 55.04 (0.92) & 13.37 (0.87) & 24.65 (1.16) \\
ResNet-RW   & 79.46 (0.55) &  92.35 (0.35)  & -  & - & - & - & - & -\\
CLIP ZS & 63.29 (0.63) & 84.47 (0.47) & 63.29 (0.63) & 84.47 (0.47) & 63.29 (0.63) & 84.47 (0.47) & 63.29 (0.63) & 84.47 (0.47) \\
CLIP LP & 86.22 (0.45) & 96.32 (0.25) & 76.41 (0.57) & 91.97 (0.36) & 70.42 (0.59) & 88.72 (0.42) & 46.99 (0.68) & 69.20 (0.62) \\
LaBo & 83.00 (0.03) & 95.17 (0.01) & 73.66 (0.07) & 90.51 (0.10) & 69.15 (0.04) & 87.27 (0.06) & 54.21 (0.06) & 76.79 (0.08) \\
CaFo & - & - & 69.48 (0.28) &  88.99 (0.24)  & 65.52 (1.21) & 86.41 (2.17) & \textbf{66.48 (0.38)} & \textbf{88.29 (0.08)} \\
FLYP & 86.41 (0.47) & 96.58 (0.28) & 78.68 (0.52) & 93.03 (0.32) & 73.11 (0.57) & 90.47 (0.38) & 50.28 (0.69) & 70.68 (0.62) \\
\textbf{GIST (Ours)} & \textbf{87.64 (0.44)} & \textbf{96.86 (0.26)} & \textbf{79.52 (0.55)} & \textbf{93.29 (0.34)} & \textbf{74.30 (0.57)} & \textbf{90.88 (0.38)} & 51.50 (0.66) & 72.88 (0.60) \\
\bottomrule
\end{tabular}
\caption{Fine-tuning with GIST captions achieves best top-1 and top-3 accuracy for CUB200-2011 in comparison to baseline methods for full-shot, 5-shot, and 3-shot regimes. We report the average accuracy and standard deviation over three k-shot samples. For the full dataset and for the CLIP zero-shot experiments, we compute statistics over 1000 bootstrapped samples of the test set.}
\label{tab:cub}
\end{table*}
 
\begin{table*}[!htbp]
\centering
\small
\renewcommand{\arraystretch}{1.3} 
\rowcolors{2}{gray!25}{white}
\begin{tabular}{*9c}
\toprule
Methods &  \multicolumn{2}{c}{Full Dataset} & \multicolumn{2}{c}{5-shot} & \multicolumn{2}{c}{3-shot} & \multicolumn{2}{c}{1-shot}\\

{}   & \textit{Top-1}   & \textit{Top-3}   & \textit{Top-1}   & \textit{Top-3} & \textit{Top-1}   & \textit{Top-3} & \textit{Top-1}   & \textit{Top-3}\\
\midrule
ResNet   & 98.54 (0.24)  &  99.88 (0.07)  & 81.35 (0.35) & 91.30 (0.30) & 72.84 (1.31) & 86.45 (0.47) & 48.47 (0.61) & 64.78 (0.28)  \\
ResNet-RW   & 98.34 (0.26) &  99.79 (0.09)  & - & - & - & - & - & -  \\
CLIP ZS & 76.64 (0.84) & 85.45 (0.70) & 76.64 (0.84) & 85.45 (0.70) & 76.64 (0.84) & 85.45 (0.70) &76.64 (0.84) & 85.45 (0.70) \\
CLIP LP & 99.36 (0.17) & \textbf{99.92 (0.06)} & 97.48 (0.32) & 99.72 (0.11) & 96.28 (0.38) & 99.63 (0.12) & 82.93 (0.73) & 94.50 (0.45) \\
LaBo & 99.26 (0.03) & 99.88 (0.05) & 96.06 (0.02) & 99.11 (0.01) & 92.75 (0.04) & 97.28 (0.23) & 80.63 (0.05) & 93.24 (0.14) \\
CaFo  & - & - & 95.40 (0.68) & 99.35 (0.15) & 93.42 (0.56) & 98.89 (0.23) & \textbf{86.50 (0.95)} & \textbf{95.98 (0.23)} \\ 
FLYP & 99.47 (0.13) & \textbf{99.92 (0.06)} & 97.80 (0.33) & 99.76 (0.10) & 96.89 (0.33) & \textbf{99.76 (0.10)} & 84.56 (0.71) & 95.03 (0.43) \\
\textbf{GIST (Ours)}  & \textbf{99.60 (0.15)} & \textbf{99.92 (0.06)} & \textbf{97.88 (0.29)} & \textbf{99.80 (0.09)} & \textbf{97.02 (0.34)} & 99.72 (0.10) & 85.63 (0.68) & 95.10 (0.43)\\
\bottomrule
\end{tabular}
\caption{Fine-tuning with GIST captions achieves best top-1 and top-3 accuracy for Flowers-102 in comparison to baseline methods for full-shot, 5-shot, and comparable accuracy for 3-shot. We report the average accuracy and standard deviation over three k-shot samples. For the full dataset and for the CLIP zero-shot experiments, we compute statistics over 1000 bootstrapped samples of the test set.}
\label{tab:flowers}
\end{table*}

\begin{table*}[!htbp]
\centering
\small
\renewcommand{\arraystretch}{1.3} 
\rowcolors{2}{gray!25}{white}
\begin{tabular}{*9c}
\toprule
Methods &  \multicolumn{2}{c}{Full Dataset} & \multicolumn{2}{c}{5-shot} & \multicolumn{2}{c}{3-shot} & \multicolumn{2}{c}{1-shot}\\

{}   & \textit{Top-1}   & \textit{Top-3}   & \textit{Top-1}   & \textit{Top-3} & \textit{Top-1}   & \textit{Top-3} & \textit{Top-1}   & \textit{Top-3}\\
\midrule
ResNet   & 69.79 (0.79) & 89.11 (0.79)
 & 22.31 (0.54) & 39.27 (1.34) & 14.22 (0.37) & 26.94 (0.84) & 7.02 (0.56) & 13.83 (0.70) \\
ResNet-RW & 69.33 (0.77) & 88.98 (0.53) & - & - & - & - & - & -  \\
CLIP ZS & 32.18 (0.83) & 61.12 (0.85) & 32.18 (0.83) & 61.12 (0.85) & 32.18 (0.83) & 61.12 (0.85) & 32.18 (0.83) & 61.12 (0.85) \\
CLIP LP & 65.43 (0.83) & 86.04 (0.60) & 45.13 (0.86) & 68.79 (0.77) & 39.51 (0.87) & 62.76 (0.86) & 26.01 (0.78) & 44.82 (0.85) \\
LaBo & 64.16 (0.15) & 85.27 (0.17) & 45.20 (0.02) & 70.33 (0.04) & 40.52 (0.03) & 65.39 (0.15) & 32.58 (0.28) & 56.35 (0.42) \\
CaFo  & - & - & 43.16 (1.15) & 70.73 (0.97) & 40.85 (1.46) & \textbf{68.44 (1.38)} & \textbf{40.05 (0.55)} & \textbf{69.78} (0.17) \\  
FLYP & 71.70 (0.75) & 90.29 (0.50) & 48.07 (0.87) & 70.00 (0.79) & 39.45 (0.86) & 63.28 (0.86) & 27.09 (0.78) & 46.87 (0.87) \\
\textbf{GIST (Ours)} & \textbf{72.27 (0.76)} & \textbf{90.42 (0.51)} & \textbf{49.44 (0.86)} & \textbf{72.93 (0.78)} & \textbf{41.84 (0.88)} & 67.30 (0.80) & 28.52 (0.78) & 48.85 (0.87)\\

\bottomrule
\end{tabular}
\caption{Fine-tuning with GIST captions achieves best top-1 and accuracy for FGVC Aircraft in comparison to baseline methods for full-shot, 5-shot, and comparable accuracy for 3-shot. We report the average accuracy and standard deviation over three k-shot samples. For the full dataset and for the CLIP zero-shot experiments, we compute statistics over 1000 bootstrapped samples of the test set.}
\label{tab:aircraft}
\end{table*}

\begin{figure*}[!htb]
\begin{center}
\centerline{\includegraphics[width=\linewidth]{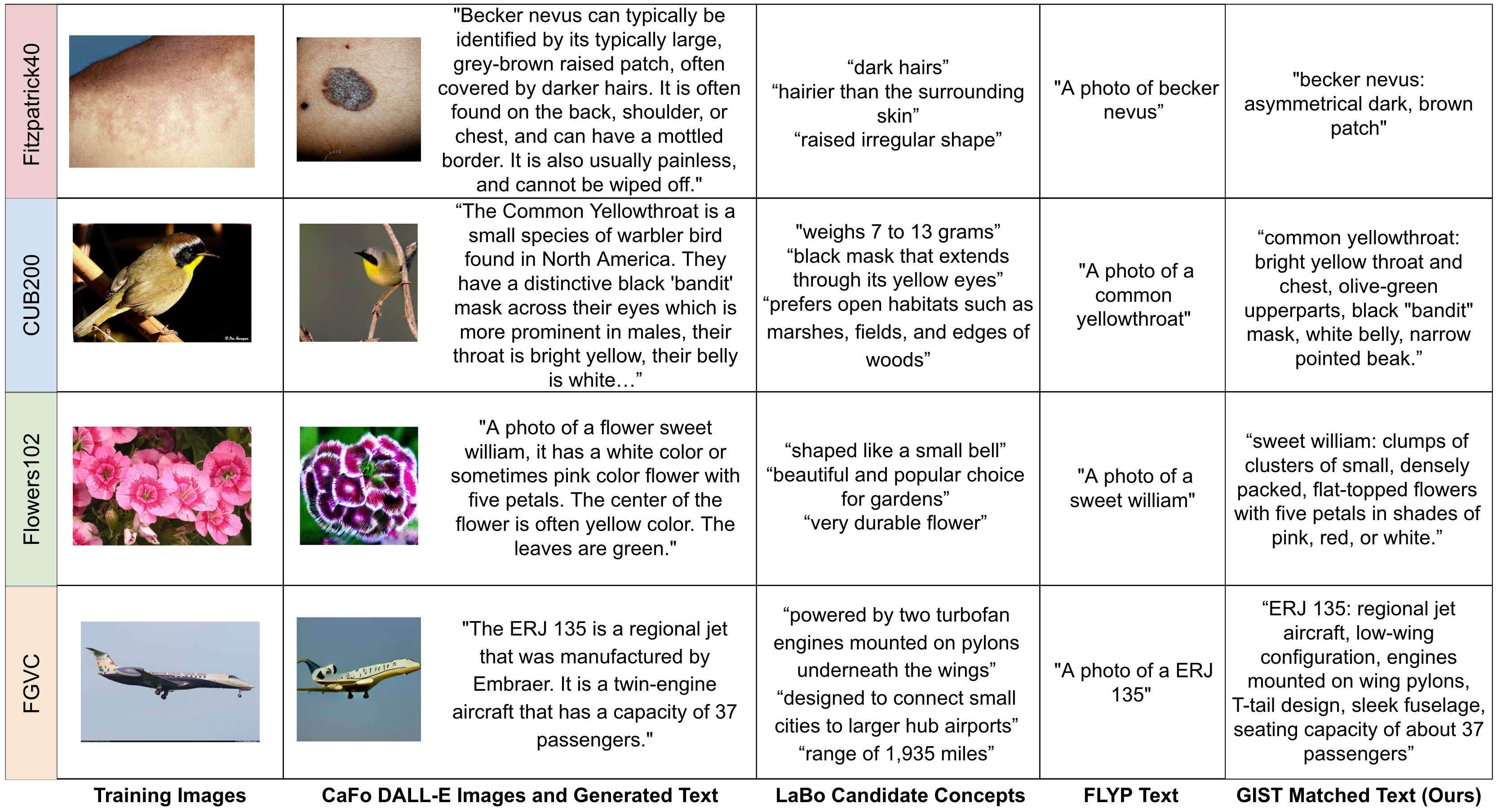}}
\caption{Qualitative examples of dataset training images and corresponding generated data from each method. Compared to other methods, GIST provides concise captions that capture key visual features in the image relevant to classification.}
\label{fig:qualitative}
\vspace{-7mm}
\end{center}
\end{figure*}

\begin{figure*}[!htb]
\begin{center}
\centerline{\includegraphics[width=\linewidth]{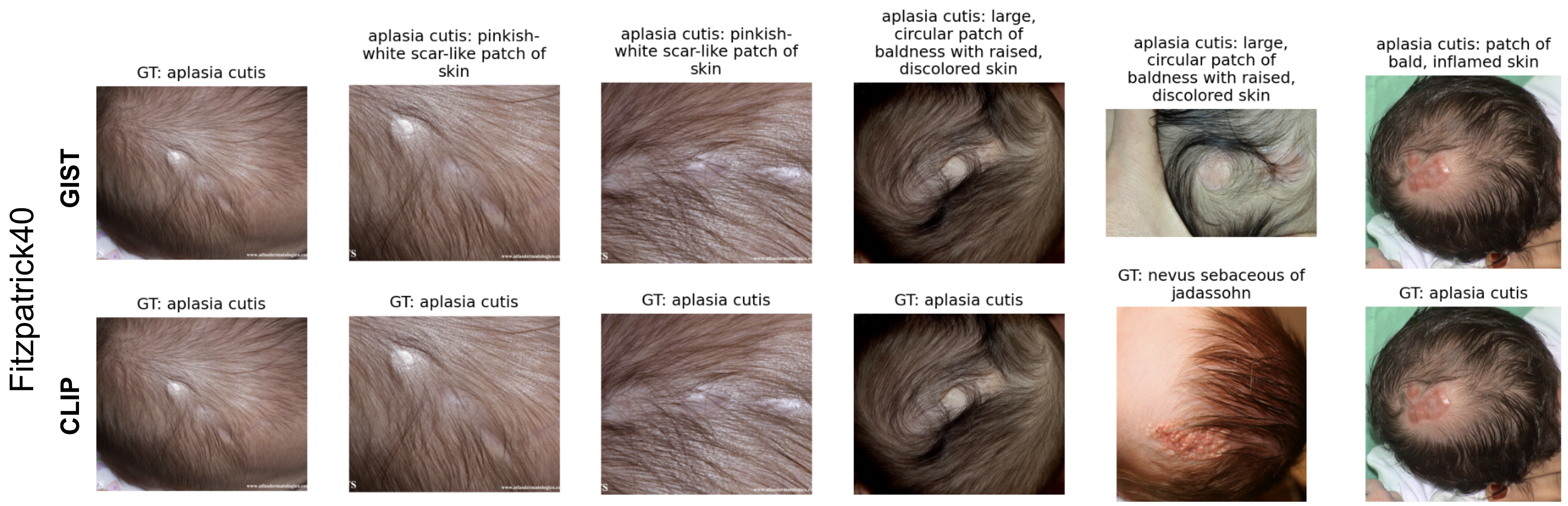}}
\caption{A qualitative example of the GIST fine-tuned representation space and the pretrained CLIP representation space for a test example that our method predicts correctly and CLIP LP predicts incorrectly. The left column shows a test image with its ground truth (GT) label. The other five columns show the five nearest neighbor training images, in descending order, of the test image with either the corresponding top matched caption (GIST) or the ground truth class name (CLIP). Our approach captures domain-specific details (e.g. ``circular patch of baldness''), whereas CLIP focuses up on image features that may not capture class differences.}
\label{fig:nn}
\vspace{-7mm}
\end{center}
\end{figure*}

\subsection{Fine-Grained Classification Results}

\paragraph{Few-Shot Classification.} Results are shown in Table~\ref{tab:fitz}, Table~\ref{tab:cub}, Table~\ref{tab:flowers}, and Table~\ref{tab:aircraft} for the Fitzpatrick, CUB200-2011, Flowers, and FGVC datasets, respectively. We report the accuracy and standard deviation on the full test set for each dataset. For each \textit{k}-shot learning experiment, we average the test results over three samples of \textit{k} training images. For the full dataset and for the CLIP zero-shot experiments, we compute statistics over 1000 bootstrapped samples of the test set. We find that GIST has better performance than all baseline methods except for the 1-shot evaluation setting. We find the CaFo performs best for the 1-shot setting. CaFo generates DALL-E \citep{DALLE} images of each class to aid in few-shot classification. These DALL-E images are relatively, but not perfectly accurate, as shown in Figure~\ref{fig:qualitative} and the supplement. We hypothesize that the relative improvement of CaFo over GIST is due to GIST overfitting in the 1-shot case and that the generated images, while imperfect, are helpful in the very-low data regime. We analyze that the Flowers102 and CUB200 dataset have relatively high image-text alignment in the pretrained CLIP embedding space as shown by the CLIP zero-shot accuracies. In contrast, the Fitzpatrick40 and FGVC datasets have much lower CLIP zero-shot accuracies, indicating they are less well represented in the pretrained CLIP. We show our method is able to achieve state-of-the-art classification performance for all four datasets despite the difference in initial image-text alignment.

\paragraph{Qualitative Results.} In addition to quantitative results, we also qualitatively examine the generated data for each method. While our method, CaFo, and LaBo all use the same GPT models, each method prompts GPT differently and has different ways of using the generated text. Furthermore, CaFo uses DALL-E to generate images for training in addition to text. Figure~\ref{fig:qualitative} shows examples of ground truth images from each dataset, corresponding generated text from each method, and generated DALL-E images (for CaFo). Our image-text matching and text summarization steps ensure that our matched text is concise and image-specific. In comparison, CaFo generates long text that, while informative, contains extraneous words and might not be specific to the described image. CaFo's DALL-E images convey correct coarse-level details, but are not high-quality. In the case of the Fitzpatrick dataset, the DALL-E images often contain class-inconsistent details such as incorrect shape, texture, or color of the disease. LaBo generates text concepts for each class, uses a selection function to narrow the set of candidate concepts, and then weights the similarity between classes and each concept. We select and show three candidate concepts per dataset and demonstrate that some concepts carry useful, accurate information, while others are generic or not descriptive of visual features. FLYP creates text by appending ``A photo of a'' to each class name. They achieve impressive results with their short captions, but do not generate interpretable descriptions. Furthermore, their captions do not capture any visual descriptors that describe and differentiate classes.  

We also show an example of learned features from our fine-tuned representation space and CLIP's pretrained representation space. Figure~\ref{fig:nn} shows a test image (left column) that our method classifies correctly and CLIP LP classifies incorrectly when trained on the full training set. The remaining columns show the nearest neighbor training images to the test image using cosine similarity over the L2 normalized embeddings. For the GIST method, we show the top matched caption for each training nearest neighbor image. For CLIP, we show the ground truth label for each image. In this dermatology example, GIST and CLIP have most of the same nearest neighbors, however, GIST corrects the fourth nearest neighbor compared to CLIP. Even though the GIST nearest neighbor in the fifth column does not visually look similar to the test image, they are close in representation because they share class-specific features, as demonstrated in the text description (e.g. ``circular patch of baldness''). The GIST fine-tuned method seems to capture domain-specific features, whereas CLIP focuses on image-specific features.

\begin{table}[htb]
\centering
\small
\renewcommand{\arraystretch}{1.3}
\rowcolors{2}{white}{gray!25}
\begin{tabular}{*9c}
\toprule
    Method & Full & 5-shot & 3-shot & 1-shot \\
    \midrule
    GIT & 74.17 (2.72) & 42.96 (2.99) & 38.32 (2.98) & 19.22 (2.32) \\
    BLIP2 & 73.83 (2.75) & 42.66 (3.03) & 37.49 (2.93) & 18.82 (2.29) \\
    GIST & \textbf{75.77 (2.67)} & \textbf{46.96 (3.13)} & \textbf{41.24 (2.95)} & \textbf{21.06 (2.38)}  \\
    \bottomrule
\end{tabular}
\caption{Fine-tuning with GIST captions outperforms finetuning with off-the-shelf GIT and BLIP2 generated captions for fine-tuning for fine-grained classification on all-shot regimes of the Fitzpatrick40 dataset.}
\label{tab:table-captioning}
\end{table}

\begin{table*}[htb]
\centering
\renewcommand{\arraystretch}{1.3}
\rowcolors{2}{white}{gray!25}
\begin{tabular}{*9c}
\toprule
\textbf{Methods} & RN50 & ViT-B/16 & ViT-B/32 & ViT-L/14 & ViT-L/14@336px \\
\midrule
    CLIP LP & 31.97 (2.85) & 38.37 (2.93) & 35.88 (2.90) & 42.27 (3.14) & 41.49 (2.99) \\
    CLIP ZS & 1.48 (0.73) & 2.98 (1.03) & 1.79 (0.81) & 1.48 (0.74) & 2.94 (1.06) \\
    LaBo & 23.16 (0.52) & 31.44 (0.26) & 29.60 (0.26) & 33.91 (0.26) & 32.17 (0.25) \\
    CaFo & 32.48 (4.40) & 37.01 (3.04) & 35.91 (2.00)  & 40.32 (1.71) & 40.32 (0.63) \\
    FLYP & 32.58 (2.84) & 40.35 (3.03) & 37.25 (2.87) & 43.97 (3.05) & 42.18 (2.98) \\
    \textbf{GIST (Ours)} & \textbf{33.40 (2.81)} & \textbf{43.91 (3.01)} & \textbf{42.08 (2.97)} & \textbf{47.45 (3.12)} & \textbf{46.96 (3.13)}\\
   \bottomrule
\end{tabular}
\caption{GIST-generated captions improves on 5-shot Fitzpatrick40 classification over a selection of different CLIP models. The top-3 GIST matched captions are used for fine-tuning. The relative improvement of the GIST-generated captions over the baselines depends on the CLIP model used for fine-tuning.}
\label{table-clip-models}
\end{table*}

\begin{table*}[htb]
\centering
\renewcommand{\arraystretch}{1.3}

\rowcolors{2}{white}{gray!25}
\begin{tabular}{*9c}
\toprule
    \textbf{Methods} & 1 & 2 & 3 & 4 & 5 \\
    \midrule
    Fitzpatrick 5-shot & 44.10 (2.97) & 44.43 (3.04) & \textbf{46.96 (3.13)} & 45.15 (3.01) & 46.69 (3.03) \\
    CUB 5-shot & \textbf{79.52 (0.55)} & 79.14 (0.55) & 79.19 (0.54) & 79.34 (0.54) & 79.25 (0.54) \\
    Flower 5-shot & \textbf{97.88 (0.29)} & 97.52 (0.31) & 97.52 (0.31) & 96.96 (0.35) & 97.06 (0.34)  \\
    Aircraft 5-shot & 49.13 (0.87) & 48.69 (0.86) & 49.33 (0.86) & \textbf{49.44 (0.86)} & 48.54 (0.86) \\
    \bottomrule
\end{tabular}
\caption{The optimal number of GIST-generated captions is dataset dependent. The optimal number of matched captions for FGVC-Aircraft is four and the optimal number for Fitzpatrick40 is three, while the optimal number of matched captions for CUB200-2011 and Flower102 is 1. We report average accuracy and standard deviation on 5-shot classification tasks for between top-1 to top-5 matched captions per image.}
\label{tab:table-captions}
\end{table*}

\begin{table}[!htbp]
\centering
\renewcommand{\arraystretch}{1.3} 
\rowcolors{2}{gray!25}{white}
\begin{tabular}{*5c}
\toprule
\textbf{Methods} &  \multicolumn{2}{c}{5-shot} \\
{}   & \textit{Top-1} &   \textit{Top-3}  \\
\midrule
FLYP & 42.18 (2.98) & 65.50 (2.96) \\
GPT-descriptions-long & 42.25 (3.12) & 66.90 (2.97)\\
GPT-descriptions-GIST & \textbf{44.10 (2.97)} & \textbf{68.20 (2.94)} \\
\bottomrule
\end{tabular}
\caption{Using shortened matched captions rather than the long original GPT-generated captions ($\approx$2 sentences) for fine-tuning helps downstream classification on the 5-shot Fitzpatrick40 dataset. For each experiment, only the top-1 caption is used to fine-tune CLIP. For all experiments, we fine-tune the ViT-L/14@336px CLIP model.}
\label{tab:caption-length}
\end{table}

\subsection{Method Analysis} \label{sec:method-analysis}
In this section, we analyze and provide insight into our method design choices. We compare GIST to off-the-shelf captioning methods and we study the impact of CLIP model, number of matched captions per image, and caption length on classification performance. We also compare against a visual grounding method and find that our CLIP fine-tuning results in much better classification accuracy.

\paragraph{Captioning Model Comparison} We compare GIST to off-the-shelf captioning models, GIT and BLIP. Table~\ref{tab:table-captioning} shows that using GIST captions to finetune outperforms traditional captioning method captions for fine-grained classification on all shots regimes of the Fitzpatrick40 dataset. For all generated captions, we append class label names. We provide qualitative examples of GIST, GIT, and BLIP captions in the supplement. 

\paragraph{CLIP Model Selection.} We show classification accuracy on the 5-shot Fitzpatrick40 dataset using different CLIP models for each method. Table \ref{table-clip-models} shows that overall classification performance for both the linear probing baseline and our method can differ significantly depending on the CLIP model. However, using GIST-generated captions for fine-tuning consistently improves upon baseline methods regardless of the CLIP model used.

\paragraph{Number of Captions.} We show the classification accuracy for using different number of captions on the Fitzpatrick40, CUB, Flower102 and FGVC 5-shot classification tasks. We evaluate using between the top 1 to top 5 matched captions for fine-tuning on 5-shot classification tasks for all datasets and report the results in Table~\ref{tab:table-captions}. We find that the best number of matched captions depends on the dataset. For Fitzpatrick40, using the top 3 captions for fine-tuning has best downstream classification. For FGVC-Aircraft, using the top 4 captions performs best. For CUB and Flower102 using the top 1 caption for fine-tuning works best. We hypothesize that datasets with lower image-text alignment in the pretrained CLIP representation space (approximated by the zero-shot accuracy) perform best with more matched captions per image because the additional information helps with aligning text concepts to images.  

\paragraph{Caption Length.} Our results show that leveraging text descriptions improves image classification, however, there are many possible design choices for how to generate and leverage text descriptions. We find that automatically shortening the original GPT generated text and then using the shortened texts to fine-tune CLIP improves downstream fine-grained classification performance over using the original longer generated text. We show results for the 5-shot Fitzpatrick40 classification experiment in Table~\ref{tab:caption-length}. 

\paragraph{Visual Grounding Comparison.} 
We compare our method to 
a visual grounding image captioning method, GIT~\cite{GIT}, for both fine-grained text generation and fine-grained image classification. 

\begin{figure}[!htb]
\begin{center}
\centerline{\includegraphics[width=\linewidth]{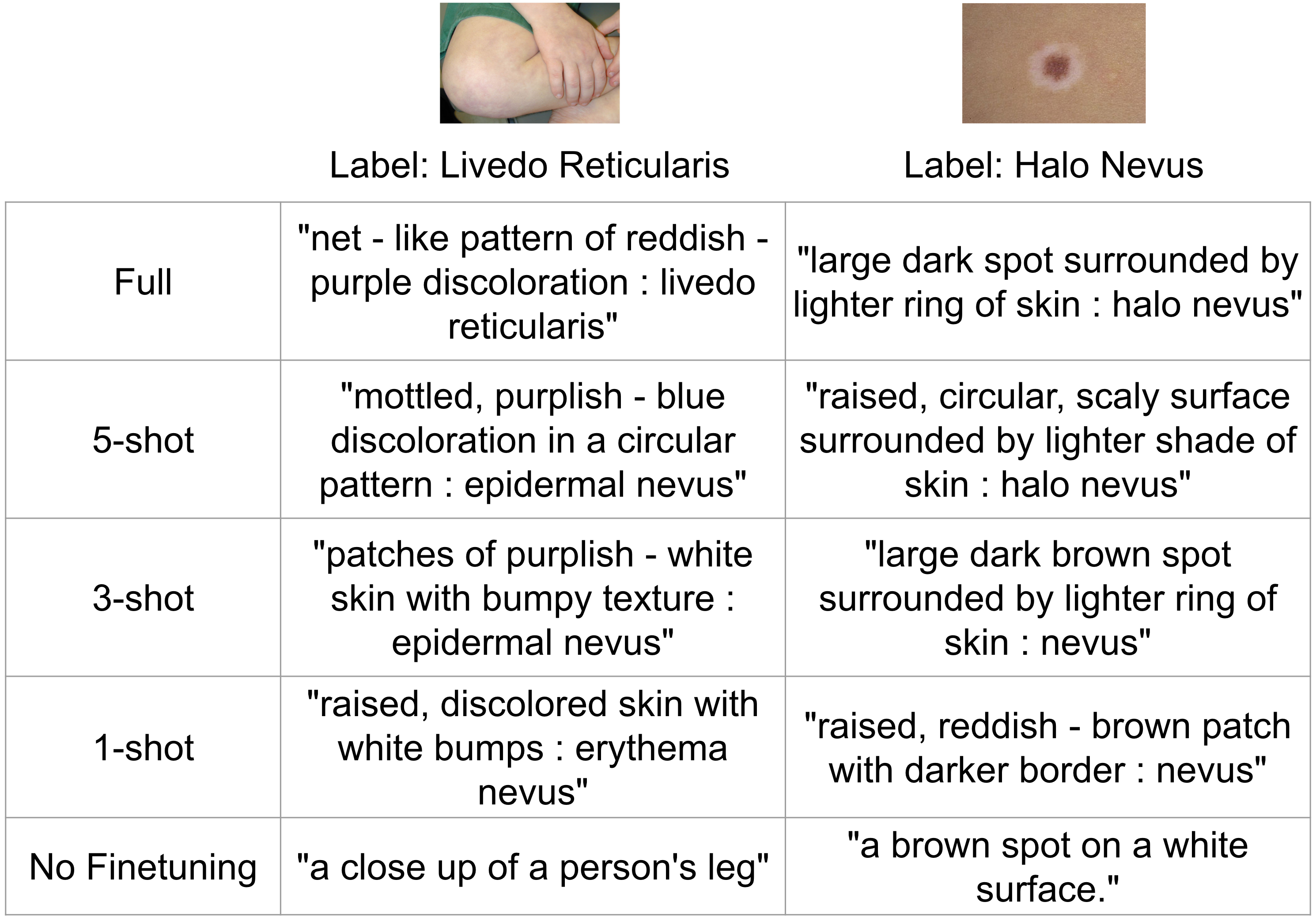}}
\caption{Examples of generated captions and class predictions from GIT for two Fitzpatrick40 test images. While the captions provide useful information and accurate predictions for the full-shot experiment, GIT overfits on the few-shot experiments. Without fine-tuning, GIT lacks fine-grained domain details.}
\label{fig:GIT-captions}
\vspace{-7mm}
\end{center}
\end{figure}

As shown in Figure~\ref{fig:placeholder} and the bottom row of Figure~\ref{fig:GIT-captions}, the pretrained GIT model outputs coarse captions that mostly focus on object-level details. In contrast, our GIST captions are able to capture more fine-grained details.

We additionally compare our fine-tuning CLIP approach with fine-tuning GIT for classification using the GIST fine-grained image-text pairs. We fine-tune GIT on our paired images and GPT-generated text descriptions with three captions per image. We append the ground truth class name to the text descriptions, in the format ``description : class name'', to form the image training captions. At inference time, we classify an image by comparing the class name in the caption to the ground truth class name. Table~\ref{tab:table-text-descriptions} shows the average accuracy and standard deviation for the full-shot and k-shot experiments on the Fitzpatrick40 dataset. Our fine-tuning with the GPT-generated matched captions outperforms fine-tuning GIT with the same image-text pairs. Figure~\ref{fig:GIT-captions} shows the GIT generated captions for two test images across different classification experiments for further analysis. Without fine-tuning, GIT focuses on objects or coarse-grained details without domain knowledge. When fine-tuning on the full training set, GIT learns to properly describe the important image features and predict the correct class name in many cases. In the \textit{k}-shot experiments, however, GIT quickly overfits to the training examples. The descriptions still contain useful features, but are attributed to the wrong class.

\begin{table}[!htb]
\centering
\renewcommand{\arraystretch}{1.3} 
\rowcolors{2}{gray!25}{white}
\begin{tabular}{*3c}
\toprule
\textbf{Experiments} & GIT & Ours
{} \\
\midrule
Full & 64.83 (2.98) & \textbf{74.59 (2.62)}\\
5-shot \textit{Top-1} & 24.02 (1.21) & \textbf{44.10 (2.97)}\\
3-shot \textit{Top-1} & 12.13 (2.40) & \textbf{40.24 (2.95)} \\
1-shot \textit{Top-1} & 1.42 (0.29) & \textbf{19.10 (2.19)}\\
\bottomrule
\end{tabular}
\caption{We report the average accuracy and standard deviation over three k-shot
samples for the Fitzpatrick40 dataset. The results show that for the same GIST captions, fine-tuning CLIP results in better classification accuracy than fine-tuning the GIT captioning method.}
\label{tab:table-text-descriptions}
\end{table}

\section{Limitations}
One current limitation of GPT is that does not always output accurate information, even with specific prompts. For our GIST approach, we found GPT-3 to work well without modification for two datasets, Fitzpatrick40 and Flowers102. The other two datasets, FGVC-Aircraft and CUB200-2011, required manual checking of the captions for rare inaccurate information for particular classes. This required a one-time upfront effort, but was not time intensive. We manually checked the captions and discarded irrelevant captions. As LLMs improve, this will not be required.

\section{Conclusion}
We present GIST as a method for generating fine-grained image-specific text descriptions and demonstrate how to use GIST to learn an aligned image-text embedding space for improved \textit{any-shot} classification. Compared to previous approaches, our text generation and image-text matching methods produce captions that are concise and specific to distinguishing image features. Our results demonstrate that fine-tuning CLIP, using contrastive learning, on our image-text pairs results in a well-aligned representation space, regardless of the original pretrained alignment. We show that applying GIST to image classification outperforms recent vision-language classification methods, black box probing methods, and image-only methods on full-shot and k-shot experiments for four diverse fine-grained datasets. 

\section{Acknowledgements}
This project was funded by the Wistron Corporation and Quanta Computer Incorporated.

\bibliography{main}

\section{Supplement}
\subsection{Fitzpatrick40 Dataset}
The original Fitzpatrick17k dataset~\cite{groh2021evaluating} has 11,788 labeled images collected from two dermatology websites, DermaAmin and Atlas Dermatologico. As mentioned in the CiRCLe paper~\cite{pakzad2023circle}, the Fitzpatrick17k dataset has many erroneous images that are unrelated to dermatology. Additionally, there is large class imbalance amongst the 114 class labels. We chose 40 of the smaller class-size labels to manually clean and form a new dataset. Fitzpatrick40 has 2,609 images, which we split into 2,115 training, 222 validation, and 272 test. There are between 25 and 73 training images per class.

The labels in our Fitzpatrick40 dataset include: becker nevus, epidermal nevus, pilar cyst, erythema nodosum, stasis edema, sun damaged skin, xeroderma pigmentosum, behcets disease, perioral dermatitis, lentigo maligna, disseminated actinic porokeratosis, halo nevus, solid cystic basal cell carcinoma, port wine stain, livedo reticularis, lichen simplex, ichthyosis vulgaris, dyshidrotic eczema, congenital nevus, naevus comedonicus, aplasia cutis, porokeratosis of mibelli, calcinosis cutis, seborrheic keratosis, erythema elevatum diutinum, mucous cyst, pilomatricoma, erythema annulare centrifigum, acrodermatitis enteropathica, pustular psoriasis, pityriasis lichenoides chronica, nevocytic nevus, lichen amyloidosis, keratosis pilaris, granuloma pyogenic, epidermolysis bullosa, drug induced pigmentary changes, basal cell carcinoma morpheiform, acanthosis nigricans, nevus sebaceous of jadassohn. 

\subsubsection{Duplicates and Non-Dermatology Images}
After noticing there were duplicate images in the full dataset, we first moved all duplicates and near duplicates in the test set to the training set. We found these duplicates by using the Sentence Transformers framework to embed the test set images into the CLIP embedding space (clip-ViT-B-32). We used cosine distance to find similar images. We found .95605 to be a good threshold to filter out duplicates in the test set. Once all duplicates were removed from the test set, we compared each remaining image to images in the validation and training set. We used the same threshold to filter out similar pairs and visually checked pairs with scores above the threshold. If we found a test set image that had near duplicates in the validation or training set, we moved both images to the training set. In addition to removing duplicates from the test set, we manually looked through the entire dataset and removed images irrelevant to dermatology (e.g. images similar to the ones mentioned in the CirCLe paper~\cite{pakzad2023circle}).

\begin{figure}[!htb]
\begin{center}
\centerline{\includegraphics[width=\linewidth]{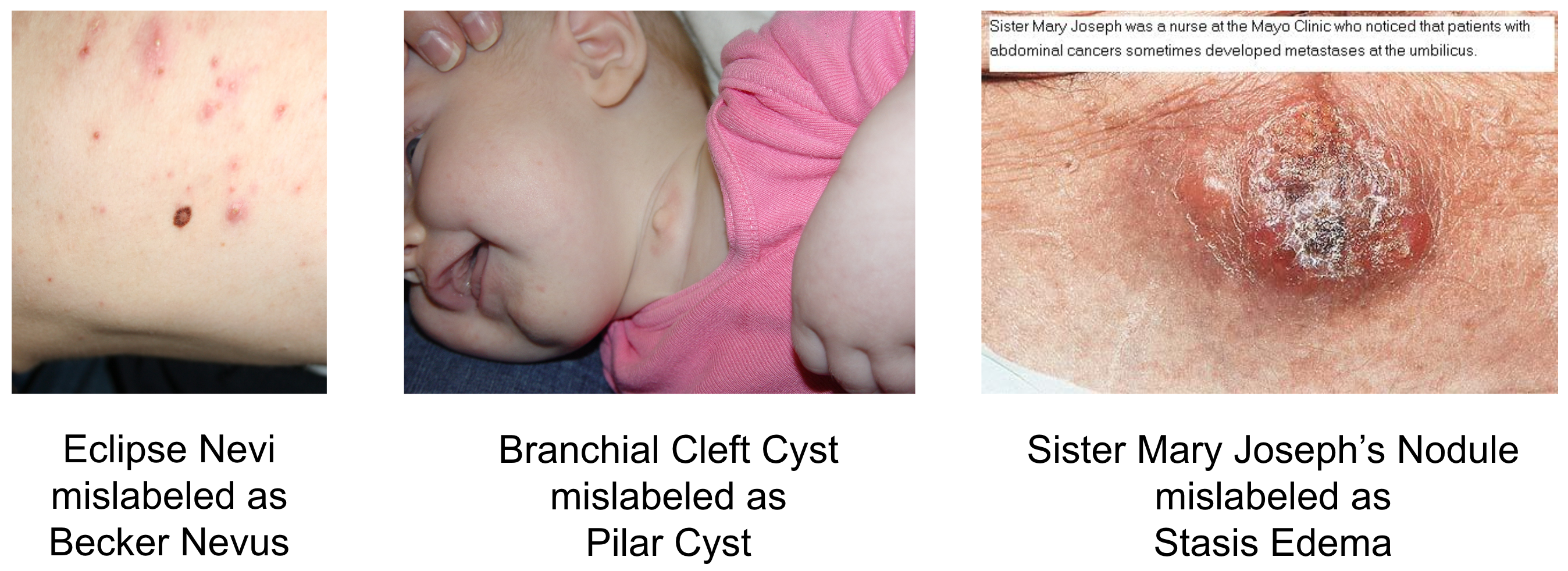}}
\caption{Examples of mislabeled images in the Fitzpatrick17k dataset. We compared the dataset label with the ground truth labels on the original source dermatology website.}
\label{fig:mislabeled-fitz}
\vspace{-7mm}
\end{center}
\end{figure}

\subsubsection{Incorrectly Labeled Images}
The Fitzpatrick dataset was collected from two dermatology websites, DermaAmin and Atlas Dermatologico. We could not find ground truth labels for the Atlas Dermatologico images. However, when working with the 40 label dataset, we realized several DermaAmin images were labeled differently in the Fitzpatrick dataset than on the website (Figure~\ref{fig:mislabeled-fitz}). We manually checked all of the DermaAmin images in the 40 label set. We looked at the DermaAmin URL label and compared it to the fitzpatrick label. If the label was different, we used Google and ChatGPT~\cite{gpt} to check whether the labels were synonyms of each other. If the Fitzpatrick label described a different disease than the DermaAmin disease, then we either 1. removed the image if the ground truth label was not one of the 40 labels in our dataset, 2. changed the image label if its ground truth label was one of the 40 labels different from its current Fitzpatrick dataset label. In some cases, the label synonyms were unclear (e.g. clinical websites differed in opinion). If we found a clinical website supporting that the labels were synonyms, we left the image as is. In future work, we would like to work with dermatologists to confirm the dataset cleaning choices.

For the training set, we found that 223 of the original 2338 images were mislabeled and the ground truth label was not one of the 40 labels. Six images were labeled as mucous cysts when they were actually sebaceous cysts (pilar cysts). Twenty images were labeled as naevus comedonicus when the source website says they are congenital nevus.

For the validation set, we found 33 of 255 images were mislabeled and the ground truth label was not one of the 40 labels. One image was labeled as mucous cysts when they were actually sebaceous cysts (pilar cysts). Three images were labeled as naevus comedonicus when the source website says they are congenital nevus. 

For the test set, we found 34 of 306 images were mislabeled and the ground truth label was not one of the 40 labels. Five images were labeled as naevus comedonicus when the source website says they are congenital nevus.

\begin{figure}[!htb]
\begin{center}
\centerline{\includegraphics[width=\linewidth]{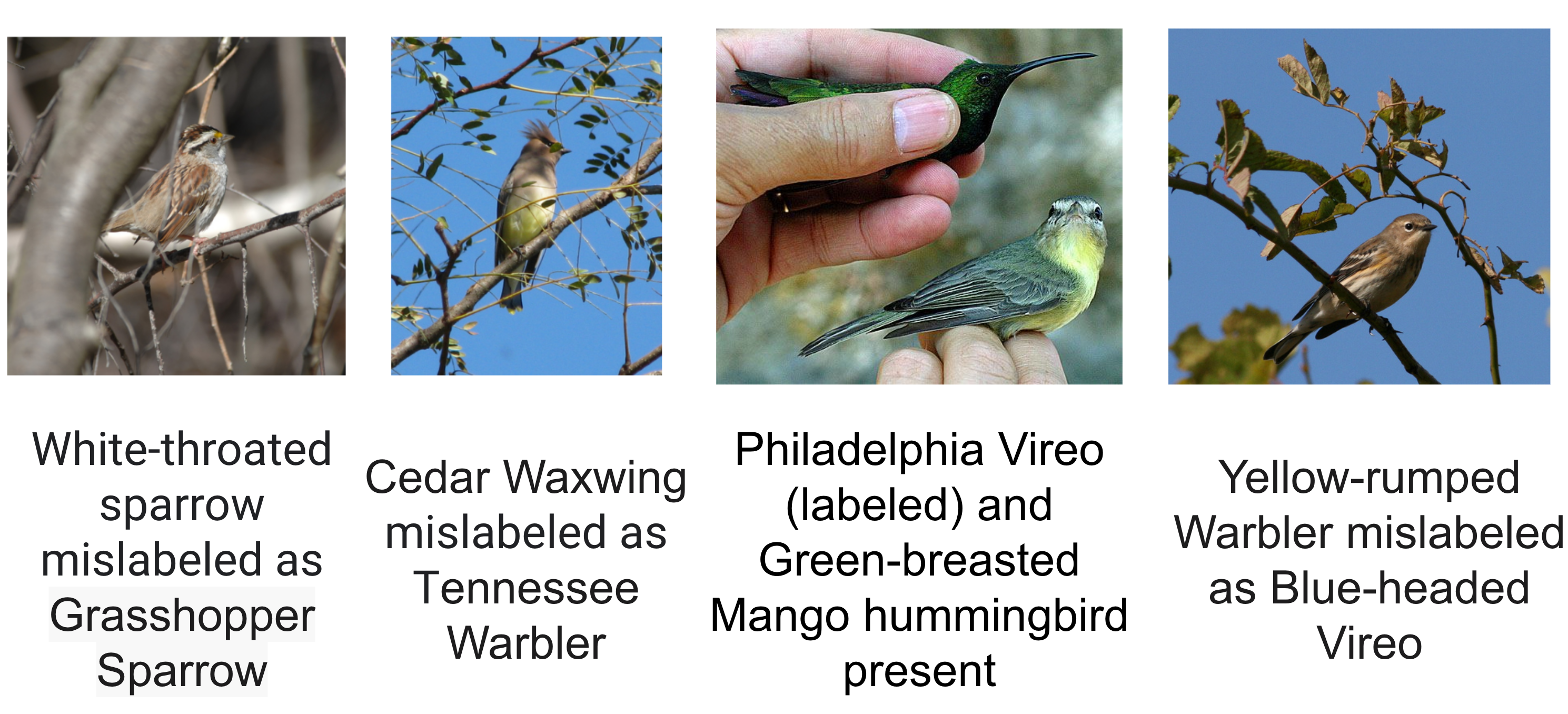}}
\caption{CUB200-2011 images that have incorrect or partially incorrect labels.}
\label{fig:birds-mislabeled}
\vspace{-7mm}
\end{center}
\end{figure}

\subsection{CUB200 Mislabeled Images}
The CUB200-2011~\cite{CUB2011} is a well-cited, commonly-used dataset for image classification. The dataset offers great diversity of images and bird species. While working with the CUB200 dataset and examining validation images misclassified by our model, we noticed that some images in the dataset are mislabeled or are partially incorrect (e.g. have two different bird species present). We show a few of these images in Figure~\ref{fig:birds-mislabeled}.

\begin{table*}[!ht]
\centering
\small
\renewcommand{\arraystretch}{1.3}
\rowcolors{2}{white}{gray!25}
\begin{tabular}{|c|c|L|L|}
\toprule
    \textbf{Class Labels} & Dataset & Long Caption & Short Caption\\
    \midrule
    Levido Reticularis & Fitzpatrick40 & ``The image shows a person's back. The skin is mottled with a network of red or purple discoloration. The discoloration appears in a reticular pattern with a lace-like structure.'' & ``mottled red or purple discoloration in reticular pattern'' \\
    Erythema Nodosum & Fitzpatrick40 & ``The image shows a person's leg with raised, red, tender lumps. The lumps are usually on the shins, and may be painful when touched. The lumps are usually red to purple in color, and may be swollen.'' & ``raised, red, tender lumps'' \\
    Red-winged Blackbird & CUB200 & ``The male Red-winged Blackbird has a black body with a red and yellow wing patch on each shoulder. It has a sharp, pointed black beak and a sleek in appearance'' & ``black body with red and yellow wing patch, sharp pointed black beak, sleek in appearance.'' \\
    Yellow-billed Cuckoo & CUB200 & ``The female Yellow-billed Cuckoo has a slender, sleek body with a long tail. She has a brownish-grey back, a white chest and belly, and prominent white spots on her dark tail. The bill is yellow on the lower mandible and black on the upper mandible.'' & ``brownish-grey back, white chest and belly, white spots on tail, yellow lower mandible, black upper mandible.'' \\
    Frangipani & Flowers102 & ``Frangipani flowers have large, showy blooms with five broad, waxy petals that create a visually captivating display. The petals can appear in various colors, including shades of white, yellow, pink, or orange with a yellow or white center. The petals often have a smooth texture.'' & ``large, waxy, and fragrant flowers that have five to nine petals in a star shape and come in a variety of colors such as white, pink, yellow, and red'' \\
    Poinsettia & Flowers102 & ``The poinsettia flower is captivating with its vibrant colors and unique structure. The bracts, which are modified leaves, are the main attraction of this flower. They are typically in shades of red, but can also be found in pink, white, or even specks of yellow. The bracts are large and oval-shaped, with a slightly wavy or scalloped edge that adds a touch of elegance.'' & ``deep red, star-shaped flowers surrounded by bright green, leaf-like bracts'' \\
    727-200 & FGVC Aircraft & ``The Boeing 727-200 is a trijet, narrow-body airliner known for its distinctive T-tail configuration. It has a relatively short and stout fuselage with three engines mounted at the rear tail section. The aircraft's wings are swept-back and positioned low on the fuselage. Its unmistakable appearance is further enhanced by the distinctive trijet engine arrangement and the characteristic dorsal intake on top of the tail.'' & ``trijet, narrow-body airliner, T-tail configuration, short and stout fuselage, three rear-mounted engines, swept-back low wings, distinctive trijet engine arrangement, dorsal intake on top of tail'' \\
    747-100 & FGVC Aircraft & ``The Boeing 747-100 is a large wide-body jetliner renowned for its iconic appearance. It features a distinctive hump-shaped upper deck, creating an instantly recognizable profile. With four engines mounted under the wings, the 747-100 has a long and elegant fuselage.'' & ``large, wide-body jetliner, hump-shaped upper deck, four engines mounted under wings, long and elegant fuselage'' \\
    \bottomrule
\end{tabular}
\caption{Examples of longer captions generated by GPT and summarized concise captions. The long captions are useful for pairing images to captions. The summarized captions provide more concise descriptions of important visual features for classification.}
\label{table-caption-length-examples}
\end{table*}

\subsection{GIST GPT Prompts}
We provide the prompts we use for each dataset to generate the long GPT captions. 

Given a set of generic body parts: face, neck, arms, torso, legs, torso, scalp, hands, feet. For the Fitzpatrick40 labels, we generate long class captions with the following prompt:

\begin{verbatim}
prompt = """You are a dermatology disease 
describer. Describe what an image of 
<<disease>> might look like on a person's 
<<body part>>."""
\end{verbatim}

For the CUB200-2011 labels, we generate long class captions for the male and female bird genders with the following prompt:

\begin{verbatim}
prompt = """You are a bird species 
describer. Describe what an image of a 
<<gender>> <<species>> might look like.
"""
\end{verbatim}

For the Flowers102 labels, we generate long class captions for flower species with the following prompt:

\begin{verbatim}
prompt = """You are a flower describer. 
Describe what an image of a flower of 
<<species>> might look like."""
\end{verbatim}

For the FGVC labels, we generate long class captions for aircraft models with the following prompt:

\begin{verbatim}
prompt = """You are an airplane model 
describer. Please describe 
distinguishing characteristics of what 
the plane looks like in 2-3 sentences.
What would a plane of type <<model>> 
look like?"""
\end{verbatim}

\subsection{GIST Caption Length} 
Our method uses GPT to generate long class-specific descriptions. After matching each training image to a description, our method uses GPT again to summarize the matched descriptions into concise captions. We show examples of the original long and the summarized short captions in Table~\ref{table-caption-length-examples}. As explained in the main paper and supported by our method analysis studies, having longer captions is useful for the image-caption matching phase. Including details, such as body part, in the longer captions increases the likelihood that matched long captions will accurately and specifically describe the images. After matching the captions and images, we find that using GPT to summarize the captions results in descriptions that contain the key visual features needed for classification.

\begin{figure}[!htb]
\begin{center}
\centerline{\includegraphics[width=\linewidth]{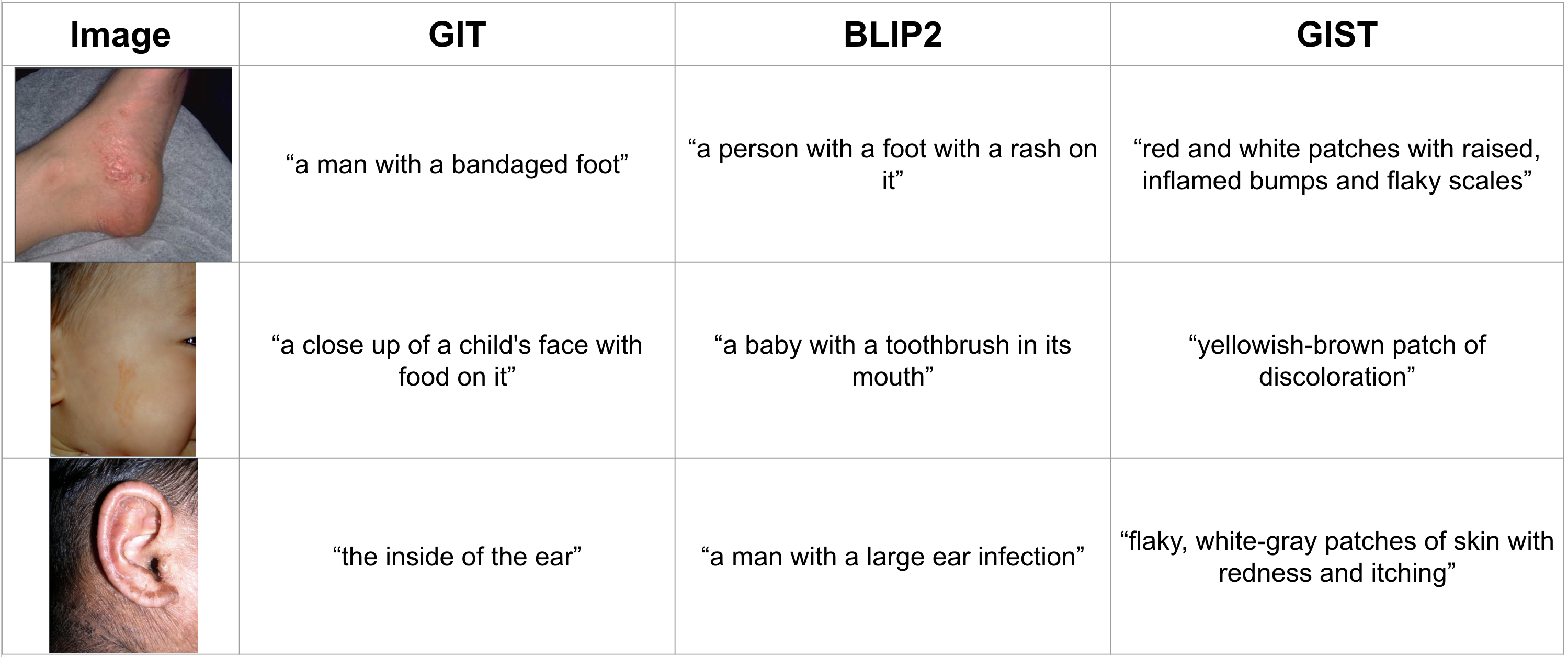}}
\caption{Example GIT, BLIP2 and GIST generated captions for Fitzpatrick40 images. For the dermatology images, GIST contains the most disease relevant information in comparison to off-the-shelf captioning models.}
\label{fig:captioning-sup}
\vspace{-7mm}
\end{center}
\end{figure}

\begin{figure}[!htb]
\begin{center}
\centerline{\includegraphics[width=\linewidth]{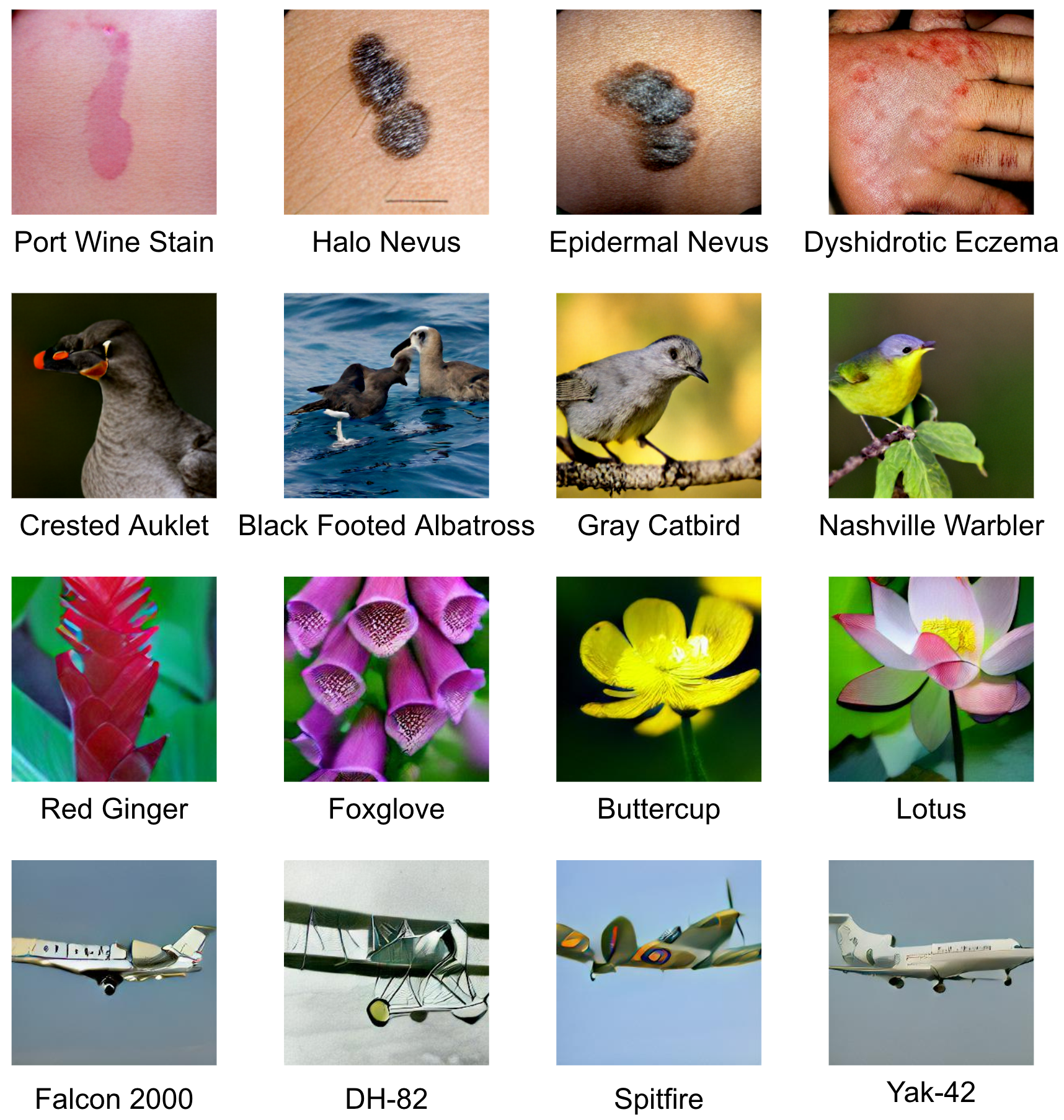}}
\caption{Example DALL-E images generated for each dataset for the CaFo method. For the dermatology images, DALL-E captures coarse-grain details, but misses key fine-grain details. For example, it generates similar images for each of the ``Nevus'' diseases, but doesn't generate the differentiating features (e.g. lighter pigmented halo for Halo Nevus). For the bird, flower, and aircraft datasets, DALL-E generates coarse and fine-grained details, but the images lack realistic qualities.}
\label{fig:DALLE-sup}
\vspace{-7mm}
\end{center}
\end{figure}

\begin{figure*}[!ht]
\begin{center}
\centerline{\includegraphics[width=\linewidth]{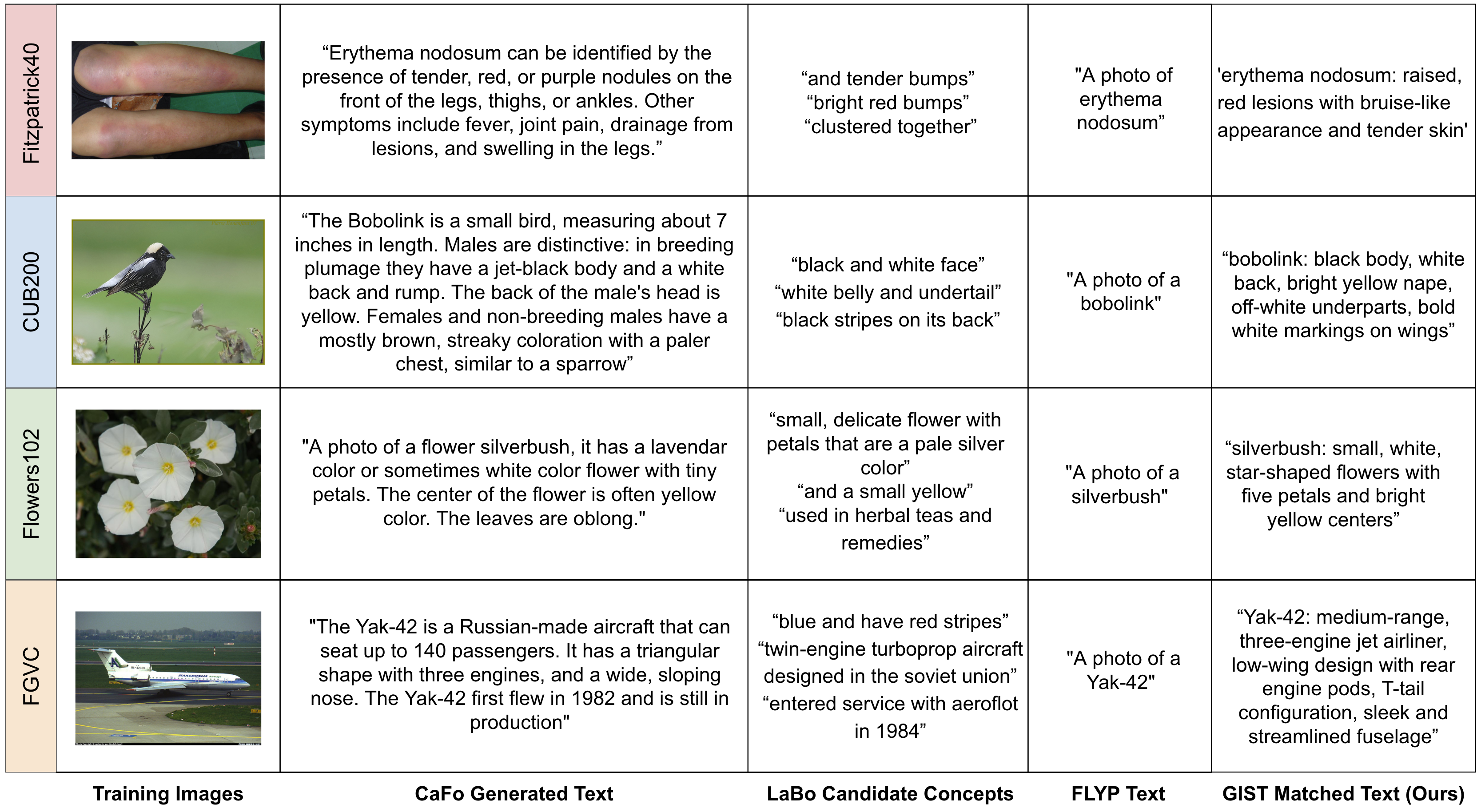}}
\caption{Generated captions and concepts from each method. Our method, GIST, generates concise but detailed captions. The baselines either generate short captions that lack visual features or generate long captions that have superfluous words. }
\label{fig:qual-results-sup}
\vspace{-7mm}
\end{center}
\end{figure*}

\subsection{Additional Qualitative Results}
We show additional generated data from each method.   

\subsubsection{Captioning Results} Figure~\ref{fig:captioning-sup} shows GIST captioning qualitative results in comparison to off-the-shelf captioning methods for three images from the Fitzpatrick40 dataset. Clearly, the GIST captions contain more disease specific captions. Some of the off-the-shelf captions are incorrect. For example, in the second row, BLIP2 identifies a toothbrush in the mouth of the baby even though neither the baby's mouth nor a toothbrush are shown.

\subsubsection{DALL-E Images} Figure~\ref{fig:DALLE-sup} shows DALL-E generated images from each dataset and their ground truth labels, which are used for the CaFo baseline method. While DALL-E is able to generate correct coarse-grain and fine-grain details for flowers, birdsm and aircrafts, the photos are not fully realistic and lack sharp details. Furthermore, it often misses important features for the dermatology dataset. For example, the halo nevus image is missing the distinctive lighter pigmented halo around the dark mole. Our results show that the DALL-E images help the CaFo method in the 1-shot classification setup, but do not boost performance compared to our method for other k-shot experiments. We hypothesize this is because the DALL-E images convey enough visual features to add information in the 1-shot setup, but lack enough realism and important features to help when there is more than one ground truth image per class.

\subsubsection{Generated Text}
We show additional examples of the generated text from each baseline method and our method in Figure~\ref{fig:qual-results-sup}.

\end{document}